\documentclass[final,3p,times]{elsarticle}

\usepackage{amssymb}
\usepackage{amsmath}
\usepackage{float}
\usepackage{booktabs}
\usepackage{xcolor}

\PassOptionsToPackage{hyphens}{url}\usepackage{hyperref}
\usepackage[absolute,overlay]{textpos}

\usepackage{background}

\backgroundsetup{
scale=10,
color=black,
opacity=0.10,
angle=45,
contents={ACCEPTED}
}

\journal{Progress in Disaster Science}

\begin{document}

\begin{frontmatter}



\title{\textsc{GraphCSVAE}: Graph Categorical Structured Variational Autoencoder for Spatiotemporal Auditing of Physical Vulnerability Towards Sustainable Post-Disaster Risk Reduction}

\author[1,2]{Joshua Dimasaka}
\ead{jtd33@cam.ac.uk}
\author[3,4]{Christian Gei{\ss}}
\author[5]{Robert Muir-Wood}
\author[1,2]{Emily So}

\affiliation[1]{
    organization={University of Cambridge},
    addressline={Department of Architecture},
    city={Cambridge},
    country={United Kingdom}}
\affiliation[2]{
    organization={Cambridge University Centre for Risk in the Built Environment}, 
    country={United Kingdom}}
\affiliation[3]{
    organization={Earth Observation Center}, 
    addressline={German Aerospace Center},
    city={We{\ss}ling},
    country={Germany}}
\affiliation[4]{
    organization={Institute of Geography}, 
    addressline={University of Bonn},
    city={Bonn},
    country={Germany}}
\affiliation[5]{
    organization={Moody's Risk Management Solutions (RMS)}, 
    city={London},
    country={United Kingdom}}

\begin{abstract}
In the aftermath of disasters, many institutions worldwide face challenges in monitoring changes in disaster risk, limiting assessment of progress towards the UN Sendai Framework for Disaster Risk Reduction 2015–2030. While numerous efforts have substantially advanced the large-scale modeling of hazard and exposure through Earth observation and data-driven methods, progress remains limited in modeling another equally important yet challenging element of the risk equation: physical vulnerability. To address this gap, we introduce \underline{Graph} \underline{C}ategorical \underline{S}tructured \underline{V}ariational \underline{A}uto\underline{e}ncoder (\textbf{\textsc{GraphCSVAE}}), a probabilistic data-driven framework for modeling physical vulnerability by integrating deep learning, graph representation, and categorical probabilistic inference, using time-series satellite-derived datasets and expert priors. We introduce a weakly supervised first-order transition matrix to capture changes in the spatiotemporal distribution of vulnerability across two disaster-affected and socioeconomically disadvantaged regions: the cyclone-impacted Khurushkul community in Bangladesh and the mudslide-affected city of Freetown in Sierra Leone. Across both case studies, the framework constructs large-scale graph representations spanning 2016–2023 and evaluates posterior compositional distributions against expert priors using Aitchison distance due to the lack of temporal groundtruth labels. The work reveals post-disaster regional dynamics in physical vulnerability, offering valuable insights into localized spatiotemporal auditing and sustainable strategies for post-disaster risk reduction.

\end{abstract}


\end{frontmatter}

\begin{textblock*}{20cm}(1.85cm,1.8cm) 
    \large \textit{\large\textit{Accepted for publication in \textit{Progress in Disaster Science} on May 20, 2026.}} 
\end{textblock*}


\section{Introduction}

In the years following a disaster, do affected communities reduce their future risk by avoiding the construction of vulnerable structures, or do they rather return to settle in pre-identified danger zones, even after experiencing their devastating impacts firsthand? While this question requires a comprehensive and multidisciplinary approach, our work offers evidence on spatiotemporal changes in physical vulnerability by applying advanced machine learning to satellite-derived datasets with prior expert belief systems. Moving beyond the scope of short-term reconnaissance missions focused on immediate recovery, our work supports large-scale disaster risk auditing, which is a crucial step towards sustainable post-disaster risk reduction and an important pillar of the 2015-2030 Sendai Framework for Disaster Risk Reduction \citep{un2015sfdrr}.

In recent years, the rise of artificial intelligence combined with the increasing availability of satellite imagery has advanced the spatiotemporal modeling of \emph{exposure} (e.g., building geometry characterization) and \emph{hazard} (e.g., deep weather forecasting models). Examples include Google Open Buildings 2.5D Temporal \citep{sirko2023high}, DLR World Settlement Footprint Evolution \citep{marconcini2021understanding}, Global Human Settlement Layer multitemporal products \citep{pesaresi2024advances}, Microsoft Aurora \citep{bodnar2025foundation}, Google GraphCast \citep{lam2023learning}, and ECMWF AIFS \citep{lang2024aifs}. However, to complete the understanding of our physical disaster risk, the element of \emph{physical vulnerability} has still remained static, limited, and coarse-grained, as reflected in the latest global assessment report \citep{undrr2025gar}. Therefore, our work advances the current state-of-practice techniques in mapping physical vulnerability using new high-resolution spatiotemporal datasets of the built environment and emerging data-driven tools to understand the fine-grained dynamics of physical vulnerability and disaster risks at large scales.

\begin{figure}[t]
    \centering
    \includegraphics[width=0.75\columnwidth]{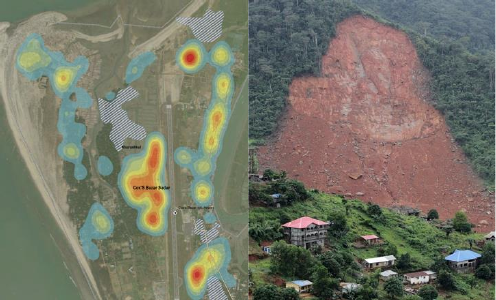}
    \caption{Devastation in (\emph{left}) the cyclone-impacted coastal community in Khurushkul, Bangladesh \citep{unitarunosat2017} and (\emph{right}) the mudslide-affected city of Freetown, Sierra Leone \citep{stedman2017}.}
    \label{fig1}
\end{figure}

\textbf{Our Contribution.} In this paper, we infer the spatiotemporal distribution of physical vulnerability in two recently disaster-stricken areas with poor socioeconomic capacities that suffered significant damage: (1) the cyclone-impacted coastal Khurushkul community in Bangladesh and (2) the mudslide-affected city of Freetown in Sierra Leone, as shown in Figure \ref{fig1}. Our work introduces \underline{Graph} \underline{C}ategorical \underline{S}tructured \underline{V}ariational \underline{A}uto\underline{e}ncoder (\textbf{\textsc{GraphCSVAE}}), a novel probabilistic data-driven framework that systematically integrates the capabilities of deep neural networks, the relational structure of graph representations, the interpretability of structured latent variables, and the probabilistic nature of categorical distributions for physical vulnerability modeling. By understanding the post-disaster regional behaviors driven by the changes in the annual distribution of physical vulnerability, our work has provided new insights into regional approaches to sustainable risk reduction.


\section{Related Work}

Previous studies on dynamic mapping of physical vulnerability have developed a variety of techniques, such as analytical Bayesian probabilistic modeling \citep{porter2014user, pittore2020variable}, cellular automata approach with Markov chains \citep{lallemant2015modeling, lallemant2017framework}, multi-agent systems based on geographic weighted regression \citep{calderon2022forecasting}, and rule-based techniques \citep{schorlemmer2020global}, which all underscore the difficulty in downscaling or disaggregating coarse-grained information into finer spatiotemporal scales given the sparsity and unavailability of building-level groundtruth labels for validation and calibration (i.e., weak supervision setting) \citep{dimasaka2024globalmappingexposurephysical}. Hence, our work leverages deep learning techniques to exploit the rich information from time-series satellite imagery and its derived products.

Furthermore, many recent advances in deep learning, particularly graph representation and variational autoencoder, have enabled the consideration of two relevant aspects: the unstructured data of building footprints and the probabilistic interpretations of the physical vulnerability categorization. Several studies applied graph representation learning in evaluating building attributes by considering the local contextual information and flexibility of data structures \citep{fill2024predicting, xu2022building, lei2024predicting, kong2024graph, dimasaka2024enhancing}. In addition, the analytical demonstration of Dirichlet-Multinomial probability distributions of \cite{pittore2020variable} to express the compositional nature of physical vulnerability, along with the categorical reparameterization trick via Gumbel-Softmax distribution introduced by \cite{jang2016categorical}, has motivated the design of our structured categorical latent representations using variational autoencoder, a deep neural network that allows probabilistic modeling by learning the parameters of assumed distributions \citep{kingma2013auto}. Together, our proposed \textbf{\textsc{GraphCSVAE}} bridges the efficient graph-structured representations of building footprints with probabilistic modeling of physical vulnerability models using deep learning on structured latent variables that follow categorical probabilistic distributions.


\section{Methodology}

This section presents the \textbf{\textsc{GraphCSVAE}} formulation, which is also referred to as the Observation Vulnerability module of \textbf{\textsc{GraphVSSM}}, a graph-based variational state-space model \citep{dimasaka2025graphvssm}. We first define the variables for exposure and physical vulnerability, followed by the description and preparation of the corresponding publicly available datasets. Then, we describe the construction of a graph-based representation and how it relates to the implementation of a variational autoencoder. Next, we detail the loss functions and metrics for evaluation with respect to our input prior information. Finally, we explain the use of a soft transition matrix to express the temporal interactions among the categories of physical vulnerability.

\subsection{Categorical Probabilistic Modeling}
\label{subsection:CategoricalProbabilisticModeling}

\begin{figure}[t!]
\centering
    \includegraphics[width=0.75\columnwidth]{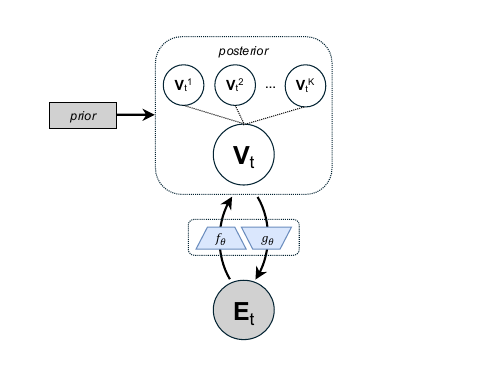}
    \caption{\textbf{\textsc{GraphCSVAE}} trains a variational autoencoder, encoder $f_{\theta}$ and decoder $g_{\theta}$, using the patterns of building exposure $E$ into a structured latent vector of physical vulnerability categories $V$ for a $posterior$ update from our input $prior$.}
    \label{fig2}
\end{figure}

In Figure \ref{fig2}, at time $t$, we define exposure $\mathbf{E}$ as an observed variable (i.e., shaded) and physical vulnerability $\mathbf{V}$ as a vector of unobserved variables with $K$ categories. Based on the probabilistic assumption of \cite{pittore2020variable}, we denote $\mathbf{V}$ as a categorical multinomial random variable:
\begin{equation}
    \mathbf{V} \sim \mathrm{Mult}(\boldsymbol{p}_{\theta}^{1}(\mathbf{E}), \ldots, \boldsymbol{p}_{\theta}^{K}(\mathbf{E}))
    \label{eq:XV}
\end{equation}
\noindent where $\boldsymbol{p}_{\theta}$ are outcomes of our encoder neural network $f_{\theta}$ with learnable parameters $\theta$. In this case, the patterns of building height from any satellite-derived products can represent $\mathbf{E}$, which can also flexibly include other relevant covariates from auxiliary datasets. 

At any location on the map, our problem defines a task of estimating the probability of observing $k^{th}$ building typology from all possible $K$ categories. Unlike their approach that uses discrete counts of buildings \cite{pittore2020variable}, we provide an alternative perspective that uses rasterized proportions of buildings instead, enabling the efficient use of geospatial data across large areal extents. 

Aligned with the catastrophe modeling, our use of a categorical probability distribution shifts the current deterministic view of the state-of-practice methods into a probabilistic interpretation of regional building exposure and physical vulnerability. When integrated with the probabilistic nature of hazard models, this approach helps key decision-makers formulate more effective, uncertainty-informed strategies across all elements of disaster risk at large scales.

\subsection{Data Preparation}

Given the present difficulty in accessing high-resolution physical vulnerability data, we used the publicly available 15-arcsecond (or 500-m) dataset of the Modelling Exposure through Earth Observation Routines (METEOR) Project \citep{huyck2019meteor}. Despite lacking temporal definition at a coarse-grained scale, we assumed that the METEOR dataset reflects the building height in 2020 of Google Open Buildings 2.5D. This limited representation of physical vulnerability serves as our prior expert belief system, which provides weak supervision to our learning objective. We preprocessed the raw METEOR dataset by normalizing the building counts for each category of physical vulnerability, as listed in Table \ref{table:vulnerability}, and resampled it to match the 50-cm spatial resolution of our building height data. The resulting normalized values become $[0,1]$, which effectively represents compositions into the probabilistic framework of the variational autoencoder through $\boldsymbol{p}_{\theta}$.

\begin{table}[H]
    \centering
    \caption{Physical vulnerability categories for each case study.}
    \label{table:vulnerability}
    \begin{tabular}{@{}l l p{0.6\linewidth}@{}}
        \toprule
        \textbf{Country} & \textbf{Label} & \textbf{Description} \\ \midrule
        Bangladesh & C3L  & Low-rise reinforced concrete\\
                   & C3M  & Same as C3L but mid-rise \\
                   & INF  & Informal constructions\\
                   & M    & Mud walls\\
                   & RS   & Rubble stone masonry\\
                   & S    & Steel\\
                   & UFB  & Unreinforced fired brick masonry\\
                   & W3   & Wood, unbraced post \& beam frame\\
                   & W5   & Wattle and daub\\ \midrule
        Sierra Leone & A    & Adobe blocks walls \\ 
                     & INF  & Informal constructions\\
                     & RS   & Rubble stone masonry\\
                     & UCB  & Concrete block, unreinforced masonry \\ 
                     & UFB  & Unreinforced fired brick masonry\\
                     & W    & Wood \\ 
                     & W5   & Wattle and daub\\
        \bottomrule
    \end{tabular}
\end{table}

In both case studies, we used the publicly available analysis-ready high-resolution 50-cm annual building height data from the Google Open Buildings 2.5D Temporal dataset, 2016-2023, which is derived from Copernicus Sentinel-2 imagery after applying a super-resolution technique \citep{sirko2023high}. It is important to note that this particular dataset serves as our proxy for $\mathbf{E}$ with the following limitations: a mean absolute error of 1.5 meters and a coefficient of determination, $R^2$, of 0.91. This mean absolute error was considered as the noise threshold for succeeding visualizations. In the respective geographic extents of post-disaster vicinities in Bangladesh and Sierra Leone, our preliminary inspection also noted that these data have no temporal gaps or missing data. Nevertheless, in this work, the spatiotemporal extent of Google Open Buildings 2.5D is adequate to demonstrate the influence of high-resolution building height patterns on the likelihood of physical vulnerability categories.

\subsection{Graph-based Representation}

Given the pre-identified geographical extent affected by disasters (i.e., as can be previewed in Figure \ref{fig5}), we divided the region into multiple non-overlapping 450-by-450 square tiles and split them into training (70\%), testing (15\%), and validation (15\%) sets with a balanced number of $\mathbf{V}$ categories. For example, the Bangladesh case study contains 1,933 tiles in total, with 1,353 tiles for training and 290 tiles each for the testing and validation sets. The Sierra Leone case study contains 1,701 tiles in total, with 1,191 tiles for training and 255 tiles each for the testing and validation sets.
 
Because of the limited number of tiles available for learning the parameters of our deep neural network in batches, instead of individual consideration of each tile, we concatenated all training square tiles and created a single undirected vulnerability graph $G^{V}_{t}=(N^{V}_{t}, A^{V}_{t}, X^{V}_{t})$ at time step $t$. $N^{V}_{t}$ is the set of nodes that represent filtered pixels with non-zero building height values. $A^{V}_{t}$ is the grid-based adjacency matrix or connectivity information between these nodes from all eight directions, which effectively forms a single training subgraph as a large, sparse binary matrix. $X^{V}_{t}$ is any feature covariates, such as our building height patterns, after applying log-normalization. However, when scaling up to larger regional extents, the dataset can instead be feasibly split by randomly sampling tiles, with each tile having its own subgraph.

Through time, $N^{V}_{t}, A^{V}_{t}, X^{V}_{t}$ may vary to indicate changes in building presence and height. We performed a similar graph construction for testing and validation sets, and prepared their corresponding individual subgraphs. To illustrate, the resulting concatenated subgraphs in the Bangladesh case study have: 7,263,547 nodes and 27,745,159 edges for training; 1,712,881 nodes and 6,553,067 edges for testing; and 1,620,374 nodes and 6,188,967 edges for the validation set. Similarly, those in the Sierra Leone case study have: 83,920,679 nodes and 162,374,972 edges for training; 19,043,847 nodes and 36,906,103 edges for testing; and 1,6726,725 nodes and 32,350,236 edges for the validation set.

\subsection{Structured Latent Variable Learning}

Using our graph-based representation, we trained a variational autoencoder network, encoder $f_{\theta}$ and decoder $g_{\theta}$, in the form of a three-layer graph convolutional neural network (GCN) \citep{kipf2016semi} with a hidden dimension of 25 each and $K$ probabilistic latent variables. In every learning iteration, both $f_{\theta}$ and $g_{\theta}$ implemented a layer-wise propagation wherein, in every layer, we propagated the update to our $G^{V}_{t}$ by following the connectivity from $A^{V}_{t}$.

The encoder network $f_{\theta}$ takes $\mathbf{E}$ as input and outputs a set of probabilistic parameters $\boldsymbol{p}_{\theta}^{1}, \ldots, \boldsymbol{p}_{\theta}^{K}$, corresponding to the $K$ categories of physical vulnerability. These parameters define a categorical latent distribution that encodes vulnerability-specific characteristics for each input instance. The decoder network $g_{\theta}$ then samples from this multinomial distribution to reconstruct $\hat{\mathbf{E}}$. As a result, the model learns a structured latent representation over the $K$ vulnerability categories, which adapts to case-specific contexts. Although this framework is transferable across regions in principle, model generalization is constrained by region-specific building typologies and local construction practices, as reflected in the distinct categories in Table \ref{table:vulnerability}. Thus, we trained separate models for each region rather than a single unifying model, as the latter would require further fine-tuning on local data or the development of shared latent representations of heterogeneous building typologies.

Due to the high computational cost for the given high-resolution 50-cm scale, we employed stochastic sampling using the combination of randomly nested subgraphs and edge dropout. To illustrate, for every training epoch, we randomly split our training subgraph into multiple nested smaller subgraphs and drop 20\% of its edges to avoid over-smoothing, which can undesirably disregard the important high values of building height. Both case studies applied the adaptive moment (Adam) optimization algorithm \citep{kingma2014adam} to update the parameters of the trained variational autoencoder network, following a step-wise learning rate schedule in which the learning rate was set to 0.01 for the first half of training epochs and reduced to 0.001 thereafter. The learning process converged after 50 epochs with 10 batches (726,354 nodes per batch) for the Bangladesh case study, and after 10 epochs with 100 batches (839,206 nodes per batch) for the Sierra Leone case study, respectively.

\subsection{Variational Learning}

As described in the preceding section, we used the categorical parameterization trick via Gumbel-Softmax distribution for a continuous, differentiable approximation, proposed by \cite{jang2016categorical}, in the sampling operation for the proportion of physical vulnerability. In symbols,
\begin{equation}
    \mathbf{V^{*}} = \frac{
        e^{(\boldsymbol{\ell_{\theta,k}} + \boldsymbol{g^{*}_k})/\tau}
    }{
        \sum_{j=1}^{K} e^{(\boldsymbol{\ell_{\theta,j}} +  \boldsymbol{g^{*}_j})/\tau}
    }
    \quad \text{for } k = 1, \dots, K
    \label{eq:ept_categorical}
\end{equation}

\noindent where $K\geq2$ for multi-category $\mathbf{V}$. $\tau$ is set to 1.0, as our scalar temperature input based on the prior distribution shape across $K$ classes, and $\boldsymbol{g^{*}}$ is sampled as:
\begin{equation}
    \boldsymbol{g^{*}} = -\log(-\log(\boldsymbol{u^{*}})), \quad \boldsymbol{u^{*}} \sim  \mathrm{Uniform}(0,1)
    \label{eq:ept_categorical_g}
\end{equation}

In addition, as introduced in Section \ref{subsection:CategoricalProbabilisticModeling}, $\boldsymbol{\ell_{\theta}}$ are logits from our trained encoder neural network.  We can further determine the corresponding probabilistic parameter $\boldsymbol{p}_{\theta}$ for the $k^{th}$ category for our multinomial probabilistic distribution by using the softmax operator, but without the stochastic sampling part and the scalar temperature input, as:
\begin{equation}
    \boldsymbol{p}^{k}_{\theta} = \frac{
        e^{\boldsymbol{\ell_{\theta,k}}}
    }{
        \sum_{j=1}^{K} e^{\boldsymbol{\ell_{\theta,j}}}
    }
\end{equation}

\subsection{Loss Function}

We trained our variational autoencoder using the sum of reconstruction loss ($\mathcal{L}^{\text{rec}}$), Kullback-Leibler divergence loss ($\mathcal{L}^{\text{KL}}$), and cross-entropy loss ($\mathcal{L}^{\text{CE}}$). At $i^{th}$ location, our encoder minimized $\mathcal{L}^{\text{KL}}$:
\begin{equation}
    \mathcal{L}_{i}^{\text{KL}}
    = \sum_{k=1}^K p_{\theta,i,k} \log \frac{p_{\theta,i,k}}{p_{0,i,k}}
\end{equation}
\noindent where $p_{\theta,i} \in \boldsymbol{p}_{\theta}$ (posterior) and $p_{0}$ represents our prior. 

Our initial findings revealed difficulty in learning diverse classifications due to weak supervision and a large number of possible categories. This limitation reflects the limited informativeness of the prior distribution, $p_{0}$, which arises from coarse-grained spatial resolution and the lack of temporal specificity across target years, resulting in insufficiently discriminative guidance for the categorical latent space under the KL regularization alone. To address this issue, we introduce an additional supervised cross-entropy loss following a semi-supervised variational learning framework \citep{kingma2014semi}, where $p_{0}$ is treated as a soft pseudo-label to directly supervise posterior predictions.
\begin{equation}
    \mathcal{L}_{i}^{\text{CE}}
    = \sum_{k=1}^K p_{0,i,k} \log(p_{\theta,i,k})
\end{equation}

Jointly, our decoder minimized $\mathcal{L}^{\text{rec}}$ using mean-squared error, wherein, for $N$ locations, $e \in \mathbf{E}$ and $\hat{e} \in \mathbf{\hat{E}}$, as:
\begin{equation}
    \mathcal{L}^{\text{rec}} = \frac{1}{N} \sum_{i=1}^{N} (e_i - \hat{e}_i)^2
\end{equation}

\subsection{Evaluation}

Evaluating the predictive accuracy and robustness of posterior distributions of physical vulnerability is a non-trivial task because of the unavailability of temporal building-level groundtruth labels. Therefore, we followed a similar metric from \cite{pittore2020variable} that expresses the compositional deviations of the resulting posterior with respect to the specified prior, indicating that any significant changes could correspond to a localized difference from the known prior, i.e., a different category of physical vulnerability. Here, we calculated the Aitchison distance, $AD$, which measures the difference between the compositions of input prior $\boldsymbol{p}_{0}$ and our posterior $\boldsymbol{p}_{\theta}$ \citep{aitchison1982statistical}.
\begin{equation}
    AD = \sqrt{ \frac{1}{2K} \sum_{i=1}^K \sum_{j=1}^K 
    \left[ \ln\left( \frac{\boldsymbol{p}_{0,i}}{\boldsymbol{p}_{0,j}} \right) - \ln\left( \frac{\boldsymbol{p}_{\theta,i}}{\boldsymbol{p}_{\theta,j}} \right) \right]^2 }
\end{equation}

It is important to note that Aitchison distance measures compositional deviation from the prior rather than predictive accuracy against groundtruth. In the absence of temporal building-level labels, this metric serves as a proxy for relative distributional change. Hence, we also performed a sensitivity analysis by varying the weights of $\mathcal{L}^{\text{KL}}$ and $\mathcal{L}^{\text{CE}}$ to examine the trade-off between prior and posterior and the influence of multiple distributions on the robustness of the inferred posterior compositions and soft transition matrices. We investigated three combinations of weights, $w_{\mathcal{L}^{\text{CE}}} / w_{\mathcal{L}^{\text{KL}}}$: $0.001, 1, 1000$. Higher values of $w_{\mathcal{L}^{\text{CE}}} / w_{\mathcal{L}^{\text{KL}}}$ (e.g., 1000) indicate a stronger weighting of the soft pseudo-labels for supervising posterior predictions, relative to the weaker informational contribution of the prior distribution.

\subsection{Soft Transition Matrix}

Because of the coarse-to-fine-grained nature of our supervised problem setting, we derived soft transition matrices, individually for each temporal step and averaged across the entire horizon, 2016-2023. Instead of using only the $k^{th}$ category with the highest posterior probability (i.e., argmax or one-hot encoding), the soft transition matrix considers the raw posterior probabilities of all $K$ categories to calculate the expected transition. 

For the average across the entire horizon $T$, at location ($x,y$) in the map with dimension ($H$, $W$), we calculated the raw $T_{ij}$ by taking the expected product of posterior probabilities $P$ between $i^{th}$ and $j^{th}$ category of physical vulnerability, from the current ($t$) to next ($t+1$) step, as:
\begin{equation}
    T_{ij} = \frac{1}{(H W)(T-1)} 
    \sum_{t=1}^{T-1}  \sum_{x=1}^{H} \sum_{y=1}^{W} 
    P_t^{(i)}(x,y)\, P_{t+1}^{(j)}(x,y)
\end{equation}

\noindent We then normalized each $T_{ij}$ so that the transition from $i^{th}$ and $j^{th}$ category sums up to one. In symbols, 
\begin{equation}
    \hat{T}_{ij} = \frac{T_{ij}}{\sum_{k=1}^K T_{ik}}
\end{equation}

We also applied this similar approach to derive a one-step soft transition matrix by setting $T$ as two consecutive temporal steps.


\section{Results and Discussion}

\begin{figure*}[htbp!]
\centering
    \includegraphics[width=\columnwidth]{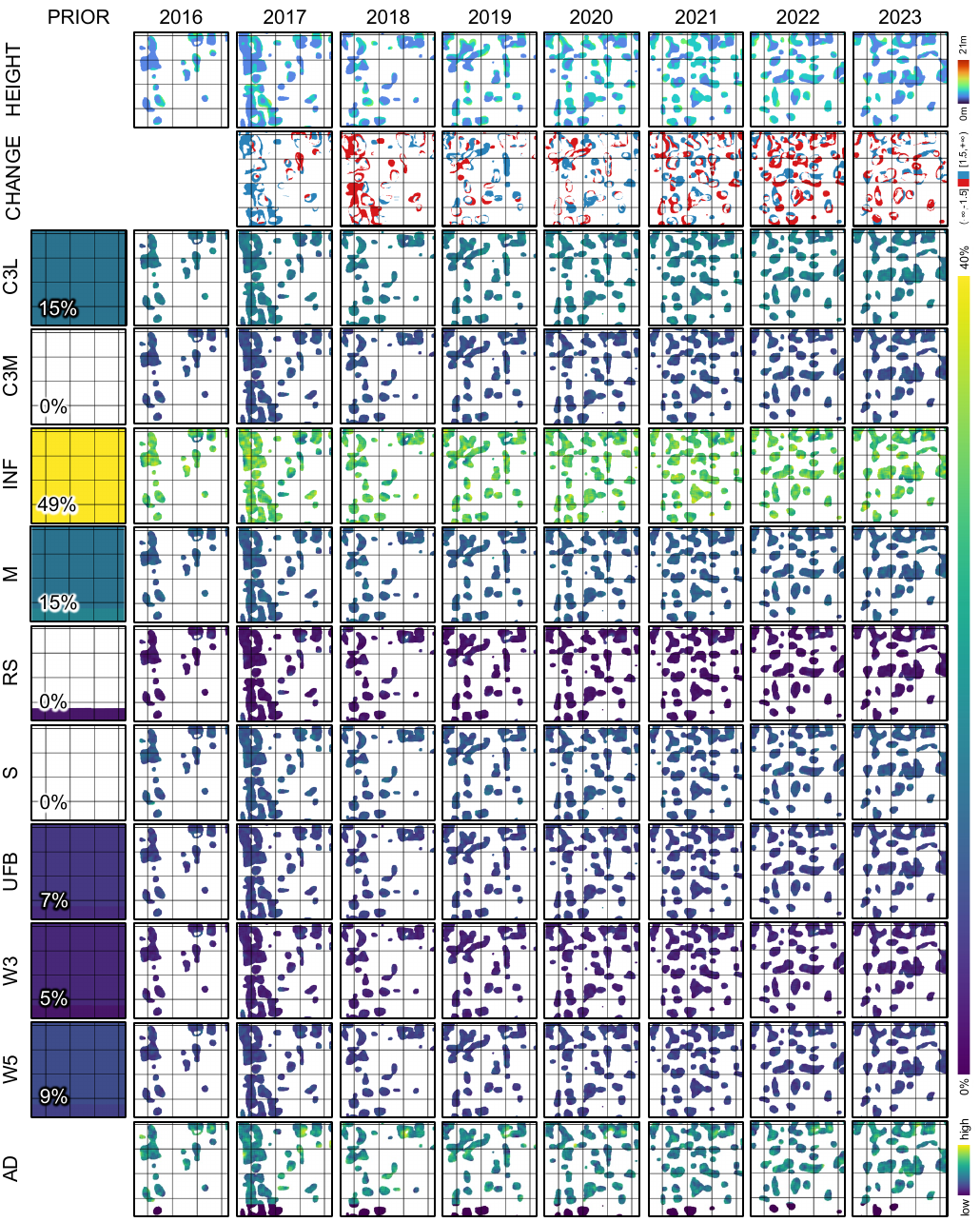}
    \caption{Prior and annual (2016-2023) posterior distribution of physical vulnerability categories in the cyclone-impacted coastal Khurushkul community in Bangladesh. The first and second top rows of subplots visualize the annual building height and its corresponding changes. From top to bottom, the next nine rows correspond to the inferred physical vulnerability. The bottom row presents the pixel-wise Aitchison distances, classified from low to high.}
    \label{fig3}
\end{figure*}

\begin{figure*}[htbp!]
\centering
    \includegraphics[width=\columnwidth]{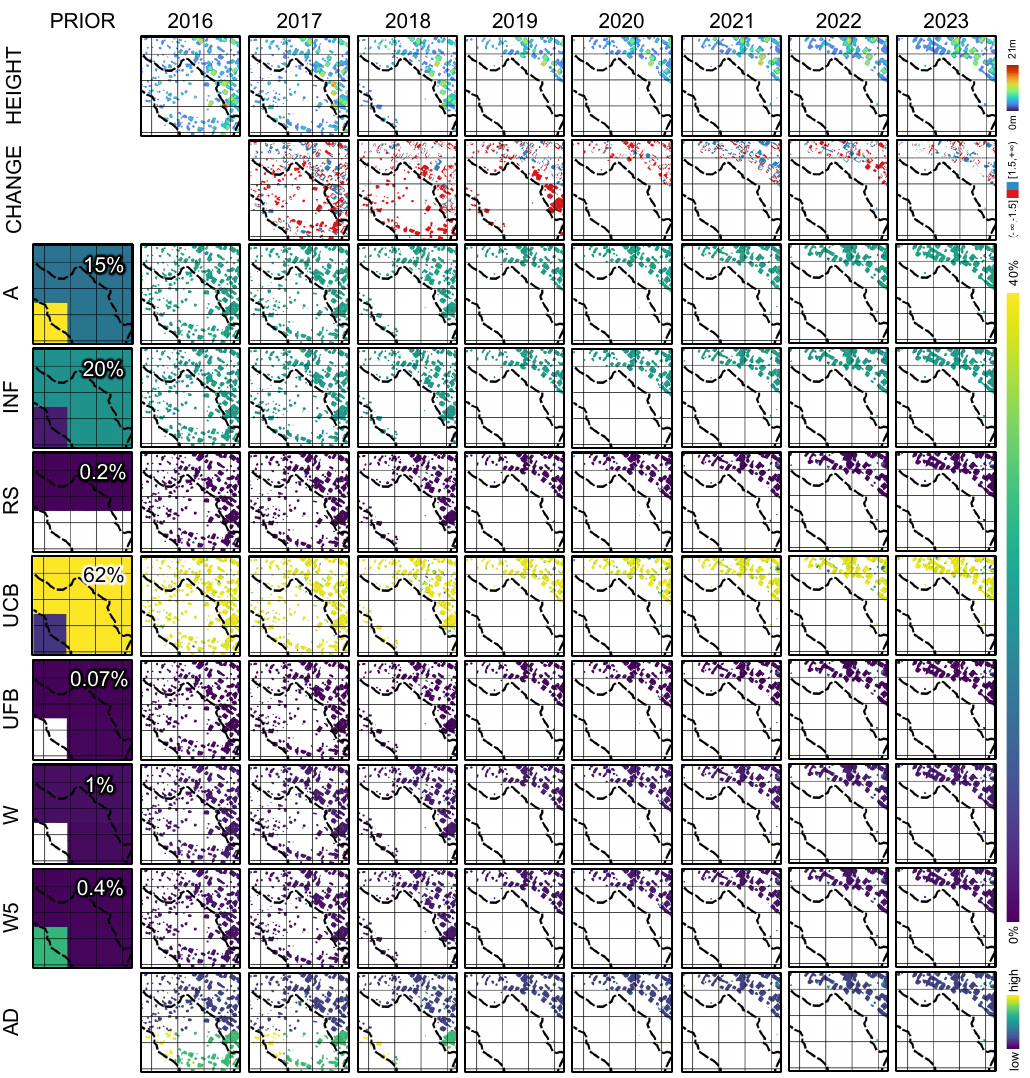}
    \caption{Prior and annual (2016-2023) posterior distribution of physical vulnerability categories in the mudslide-affected Freetown in Sierra Leone. The first and second top rows of subplots visualize the annual building height and its corresponding changes. From top to bottom, the next seven rows correspond to the inferred physical vulnerability. The bottom row presents the pixel-wise Aitchison distances, classified from low to high. The black dashed lines describe the extent of the mudslide.}
    \label{fig4}
\end{figure*}

In this section, we discuss three key insightful results on how \textbf{\textsc{GraphCSVAE}} (1) enables the probabilistic mapping of post-disaster distributions of physical vulnerability categories for two case studies in Figures \ref{fig3} and \ref{fig4}; (2) leverages temporal building exposure data in revealing the annual changes in compositions in Figure \ref{fig5}; and (3) provides a weakly supervised first-order transition matrix among the categories of physical vulnerability in Figure \ref{fig6} to facilitate the spatiotemporal audit of disaster risk.

\subsection{Mapping Post-Disaster Physical Vulnerability}


In both case studies, the high-resolution information of building presence and height from the Google Open Buildings 2.5D Temporal dataset has extended our understanding of post-disaster changes by uncovering regional community behaviors on an annual basis. For instance, considering 2016-17 as the most likely representation of the pre-disaster period, the first two top rows of Figures \ref{fig3} and \ref{fig4}, respectively, confirm the negative impacts of cyclone Mora in Bangladesh and mudslides in Sierra Leone, as indicated by the increased red signals (i.e., over 1.5-m difference) in the 2017-18 period on the second box of the second top row. Our visualization used a threshold of $\pm 1.5$ meters to highlight significant changes in building height patterns due to the prevalence of low-magnitude noises (i.e., below 1.5-m difference), which can be potentially misleading when conducting causal attribution for signs of damage or recovery. Nevertheless, despite the complex patterns of post-disaster changes, our demonstration of temporally defined building characteristics reinforces the significance of regular updating of building stock, particularly in low-capacity areas over the long term, and enables advanced monitoring using near-real-time, satellite-derived information of the built environment.

In addition, Figure \ref{fig3} shows evidence of recovery in Bangladesh in the succeeding years, but followed by negative changes, most notably in 2022, as a possible outcome of a recent flooding incident in July 2021 \citep{unitarunosat2021}. In contrast, Figure \ref{fig4} reported that the affected region in Sierra Leone remained uninhabitable for the next six years. While the case of Bangladesh reveals a community behavior of resettlement in affected areas with persisting flooding risk, the case of Sierra Leone shows that it took roughly two years for complete evacuation within and around the affected area, followed by possible continued developments in the surrounding perimeter. These two cases with different hazards illustrate how the nature, extent, and intensity of impact influence the post-disaster community behavior of recovery and reduction of future risk.


Furthermore, both Bangladesh and Sierra Leone indicate that the dominant prior categories are already highly vulnerable, which are, respectively, significantly composed of \textbf{INF}  and \textbf{UCB}, followed by the minor compositions of \textbf{C3L}, \textbf{M}, and \textbf{A}. As expected, our resulting posterior distributions consistently follow the prior composition, which inherently reflects the bias introduced by the weak supervision of coarse-grained prior information. Despite the lack of building-level labels that could have provided additional discriminative capability at the pixel level, our findings have demonstrated the value of the graphical rasterized information of fine-grained building data in learning the influence of the pixel-wise variation of building height, thereby providing weakly supervised first-order posterior distributions. Higher-order analyses can further calibrate our work with building-level validation on changes in physical vulnerability, when available. 

Compared to other similar weakly supervised efforts that used deterministic proportion-constrained approaches \citep{dimasaka2025deepc4, geiss2023benefits}, our work introduces a learnable deep inductive relationship between pixel-wise building height and physical vulnerability categories in a graph-based probabilistic framework. To illustrate further, upon closer inspection, the posterior maps show variation in probability at the pixel level since our approach takes every pixel as a node in our graph-based representation learning and incorporates its meaningful local contextual information. Our results also infer posterior values for pixels that do not have prior information, which confirms our model capability in gathering new insights from input covariates, instead of fully relying on the prior information alone.


The corresponding maps of the calculated pixel-wise Aitchison distance (AD) have demonstrated the trade-off between the \emph{exploration} of learned induced information from the patterns of building height and the \emph{exploitation} of existing prior information. In both case studies, our results consistently yield high AD, particularly for areas without complete prior information across all $K$ categories. Similar to \cite{pittore2020variable}, compared to areas with low AD, the posterior distribution with high AD observes more compositional difference from the prior, which is indicative of the greater influence induced by the patterns of building height than that of prior information. This can also be partly attributed to the limitation of poor resolution of prior information, wherein it coarsely disregards some categories.

Although our work has provided a weakly supervised first-order distribution of posterior probabilities of physical vulnerability, our approach has synthesized information on building heights in substantially improving the existing static METEOR data with finer-grained and dynamic information on the square footage of each vulnerability category. Further developments could advance understanding of regional disaster risk dynamics beyond the isolated perspectives of urbanizing exposure or intensifying hazards alone. In particular, the observed post-disaster vulnerability dynamics highlight the need for regular local monitoring through continuous mapping of building typology, enabling policymakers to identify critical areas for targeted interventions towards sustainable recovery and risk reduction.


\subsection{Examining Temporal Compositional Changes}

\begin{figure*}[t]
    \centering
    \includegraphics[width=\columnwidth]{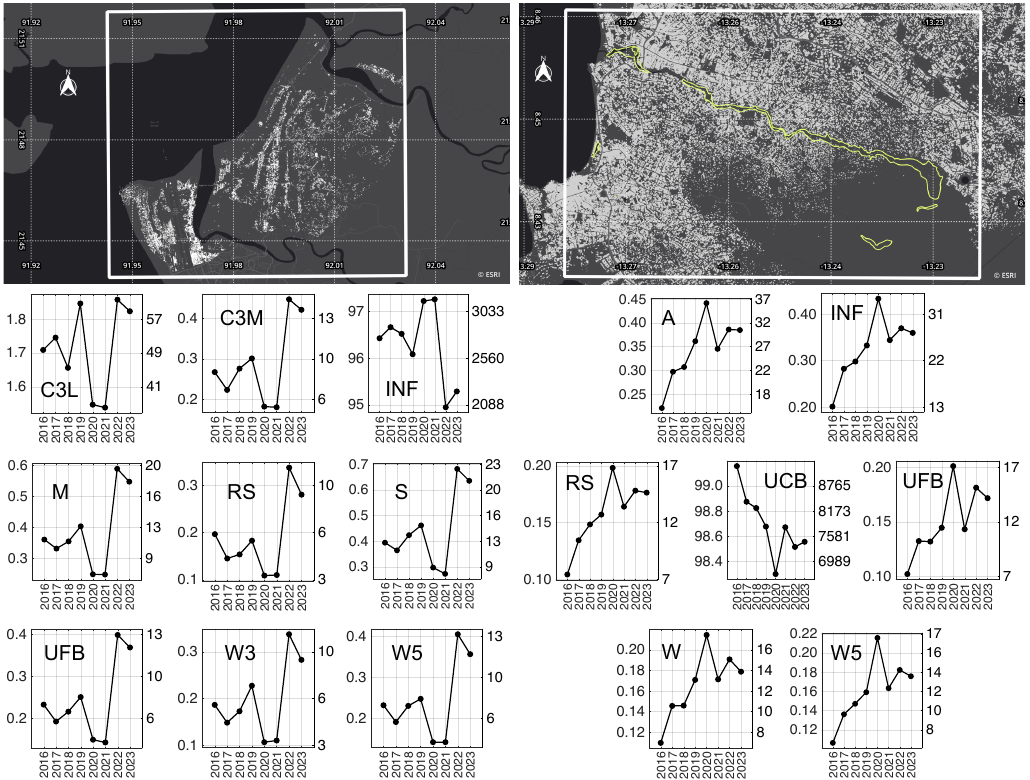}
    \caption{Annual trend of the regional mean posterior probability of physical vulnerability in (\emph{left}) the cyclone-impacted coastal Khurushkul community, Bangladesh, and (\emph{right}) the mudslide-affected Freetown, Sierra Leone. The white rectangles define the geographical extent for the calculation of the regional mean. The yellow polygon describes the affected extent of the mudslide.}
    \label{fig5}
\end{figure*}

In Figure \ref{fig5}, both case studies also exhibit one-category dominance wherein \textbf{INF} and \textbf{UCB}, respectively, allocated about 3-5\% and 1-2\% for other categories in our areas of interest in Bangladesh and Sierra Leone. This commonly observed compositional characteristic indicates that any temporal changes in the regional mean posterior probabilities are most likely influenced by the associated changes in the binary presence of input building exposure data, followed by the inferred variation based on the building height patterns. Unlike other studies that deal with spatiotemporal prediction tasks with more balanced labels, our findings underscore this uniquely underexplored challenge brought about by the complex and shared temporal nature of unbalanced categories.

Instead of examining individual minority categories, our findings effectively provide comparative analyses between the dominant and the group of non-dominant categories. For instance, Figure \ref{fig5} (\emph{left}) shows three evident temporal trends: 2017-18, 2019-20, and 2021-22. The 2017 cyclone disaster clearly caused the observed downward trend of \textbf{INF} in 2017-18, but later increased in 2019-20, indicative of a regional behavior towards rebuilding of informal constructions. The following downward trend in 2021-22, which is roughly a 2\%-decline, confirms the large displacement caused by the airport construction \citep{khan4996026tropical}. On the other hand, in Figure \ref{fig5} (\emph{right}), we observe an upward trend across all categories except \textbf{UCB}. We also report two downward trends in \textbf{UCB} in 2016-19 and 2019-20, respectively, caused by the impacts of mudslides and another flooding \citep{acaps2019}, followed by a slight reconstruction in the existing neighborhoods of \textbf{UCB} in 2020-23.

The examination of these temporal compositional changes enables comparisons of regional shifts in physical vulnerability, providing a basis for assessing the impacts of urban development and recurring hazards for local development planning. However, before these methods can be ethically deployed in already vulnerable communities, it is important to note that minority categories remain challenging and inconclusive in our study, due to the combined effects of signal noise and potentially learned relationships with building height patterns. More broadly, our work reveals important regional temporal trends at large scales and compares the impacts of two successive disasters. In Sierra Leone, the results suggest that the flooding incident caused more widespread damage than the mudslides, which are limited to mostly high-slope terrain and along the river systems. In Bangladesh, we further differentiate the quantified impacts of the 2017 cyclone on \textbf{INF} buildings from the more permanent displacement associated with airport development.

\subsection{Towards Spatiotemporal Risk Auditing}

\begin{figure*}[htpb!]
    \centering
    \includegraphics[width=\columnwidth]{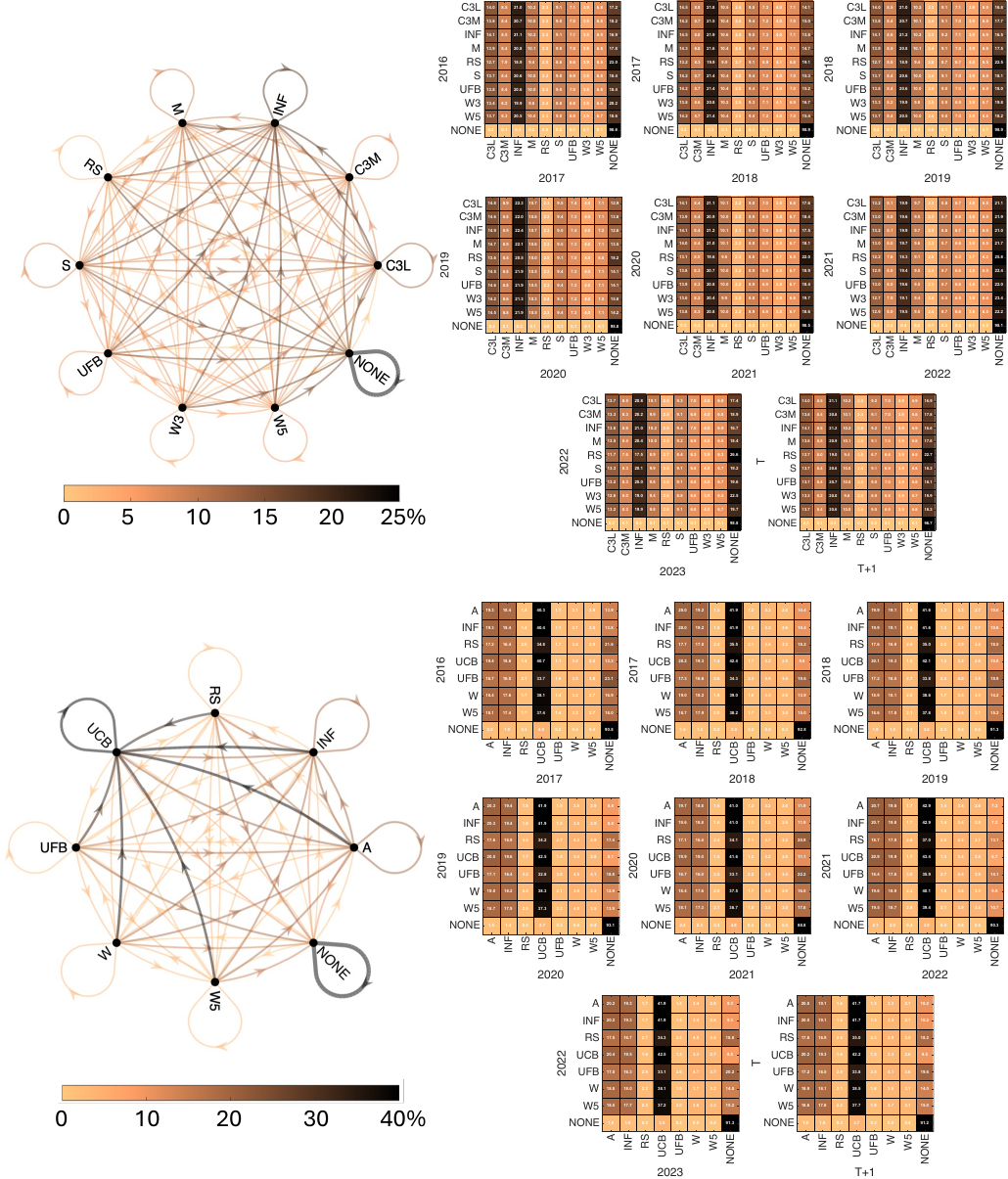}
    \caption{Graphical and tabular illustrations of the first-order transition matrices among the physical vulnerability categories in (\emph{top half}) the cyclone-impacted coastal Khurushkul community, Bangladesh, and (\emph{bottom half}) the mudslide-affected city of Freetown, Sierra Leone. The arrows signify the direction of changes, including a self-loop (i.e., retaining the existing category). All matrices show one-step transition, except the last matrix (i.e., $T$ and $T+1$), which takes the average across all temporal periods, 2016-2023.}
    \label{fig6}
\end{figure*}

Another key contribution of our work is the derivation of weakly supervised first-order \emph{transition matrices} in Figure \ref{fig6} that quantify how the `probability' of a particular category changes to another, thereby paving the way to geospatial calibration and significant extension of previous efforts, such as the Markov chain approach \citep{lallemant2015modeling} and purely statistical techniques \citep{porter2014user, pittore2020variable}, in modeling future physical vulnerability characteristics regionally. Taking advantage of the flexibility and expressivity of graph deep learning, we have established a framework that can incorporate diverse covariates from big geospatial data to influence the likelihood of physical vulnerability and its transition behavior. 

In our graphical illustrations in Figure \ref{fig6}, we observe a higher probability of self-loop transition for \textbf{NONE} and the dominant categories of \textbf{INF} and \textbf{UCB}, which confirms the gradual urban development rather than rapid construction or demolition. As anticipated, the majority of the probabilities of transition follow the limitation of the weakly supervised context of coarse-grained labels, which results in uniform transitions across categories and the absence of diagonal dominance in matrices. Correspondingly, these findings reflect the major limitations brought about by the lack of temporal definition and the coarse-grained representation of the METEOR dataset as a prior, which induces bias towards the prior in the derivation of first-order transition matrices.

In addition, the sensitivity analysis showed negligible changes in the average weakly supervised first-order transition matrix across all temporal periods (2016–2023), with the dominant transition patterns in Figure \ref{fig6} remaining largely consistent across different weight combinations. This behavior is expected because both the Kullback-Leibler divergence loss ($\mathcal{L}^{\text{KL}}$) and cross-entropy loss ($\mathcal{L}^{\text{CE}}$) are primarily dictated by the same prior information. The results also suggest that pseudo-label supervision contributes to the robustness of the variational autoencoder in learning spatiotemporal posterior distributions. Unlike deterministic proportion-constrained approaches \citep{dimasaka2025deepc4}, our framework provides a more flexible balance between prior information and newly learned representations in the estimation of transition probabilities.

Therefore, our framework enables an opportunity to extend weakly supervised first-order transition matrices to higher-order formulations by replacing probability-based representations with one-hot encoding when building-level labels are incorporated. This extension would support finer-grained probabilistic updating and more rigorous auditing of spatiotemporal distributions of physical vulnerability and disaster risk. Generalizable across hazard types, these regional transition matrices may also serve as localized indicators for cross-regional comparison, with potential applications in multiple sectors, including national policy development for vulnerable building typologies and urban planning efforts aimed at future-proofing housing systems.

\section{Conclusion and Future Work}
Reflecting on our question introduced in the beginning, our work revealed spatiotemporal findings on the post-disaster regional behavior of settlements in Bangladesh and Sierra Leone that are continually facing disaster risks. Despite the persisting challenges from the lack of building-level calibration, our proposed \textbf{\textsc{GraphCSVAE}} unified deep learning, graph representation, and categorical probabilistic inference by leveraging time-series satellite-derived datasets and coarse-grained prior information to derive a weakly supervised first-order transition matrix, which can potentially serve as a framework in auditing the regional distribution of physical vulnerability and disaster risk towards the UN Sendai Framework for Disaster Risk Reduction 2015–2030.

Importantly, the evaluation based on Aitchison distance provided a compositional perspective on changes in posterior distributions of physical vulnerability, enabling comparison against prior expert belief systems under weak supervision. While the Aitchison distance provided an effective measure of compositional change with respect to the prior, it cannot substitute for groundtruth-based accuracy assessment, underscoring the need for building-level validation data. This also highlighted a broader methodological gap in current large-scale vulnerability modeling, where localized evaluation of the fine-grained dynamics of physical vulnerability remains difficult at the building level. 

Despite the contributions of this work, several limitations remain. The unavailability of temporal building-level groundtruth labels prevents direct validation of posterior accuracy beyond compositional deviation analysis using Aitchison distance. In addition, the coarse-grained spatial resolution and lack of temporal specificity of the METEOR dataset as a prior introduce bias that propagates into both the posterior distributions and transition matrices. While the framework can flexibly incorporate diverse covariates from big geospatial data, the interpretability of learned graph representations in characterizing changes in physical vulnerability remains an open challenge. Before these methods can be ethically deployed in already vulnerable communities, careful local contextualization at the building scale is necessary, beyond the regional summary from transition matrices. Therefore, to enable more rigorous model validation, we recommend the incorporation of building-level information, even sparse or incomplete, for future work in achieving more accurate higher-order analyses.

\section*{CRediT Authorship Contribution Statement}

\textbf{Joshua Dimasaka:} Conceptualization, Data curation, Formal analysis, Investigation, Methodology, Software, Validation, Visualization, Writing – original draft, Writing – review and editing. \textbf{Christian Gei{\ss}:} Conceptualization, Project administration, Supervision, Writing – review and editing. \textbf{Robert Muir-Wood:} Conceptualization, Supervision, Writing – review and editing. \textbf{Emily So:} Conceptualization, Supervision, Funding acquisition, Project administration, Resources, Writing – review and editing.

\section*{Acknowledgments}
This work is funded by the UKRI Centre for Doctoral Training in Application of Artificial Intelligence to the study of Environmental Risks (AI4ER) (EP/S022961/1).

\section*{Data and Code Availability}
The data and code are available at \url{https://doi.org/10.5281/zenodo.16656471} \citep{dimasaka_2025_16656471} and \url{https://github.com/riskaudit/GraphCSVAE} \citep{dimasakaGitHubGCSVAE}, respectively.

\section*{Declaration of Interests}
The authors declare that they have no known competing financial interests or personal relationships that could have appeared to influence the work reported in this paper.

\bibliographystyle{elsarticle-harv} 
\bibliography{bibliography}

\end{document}